\documentclass{article}

\usepackage{arxiv}

\usepackage[utf8]{inputenc} 
\usepackage[T1]{fontenc}    
\usepackage{hyperref}       
\usepackage{url}            
\usepackage{booktabs}       
\usepackage{amsfonts}       
\usepackage{nicefrac}       
\usepackage{microtype}      
\usepackage{lipsum}
\usepackage{svg}
\usepackage{graphicx}
\graphicspath{ {./images/} }
\usepackage{amsmath} 
\usepackage{pdflscape}
\usepackage{xcolor}
\usepackage[normalem]{ulem}
\usepackage{comment}
\usepackage{longtable}
\usepackage{tabularx}
\usepackage{caption} 
\usepackage{subcaption}
\usepackage{longtable}
\usepackage{booktabs}
\usepackage[utf8]{inputenc}
\usepackage{changepage}

\usepackage{array}
\usepackage{float}
\newlength{\extralength}
\newlength{\fulllength}
\newcolumntype{C}{>{\centering\arraybackslash}X}

\title{YOLOv4: A Breakthrough in Real-Time Object Detection}

\author{Athulya Sundaresan Geetha\\[1ex]
\begin{minipage}[t]{0.90\textwidth}
\centering
\scriptsize Department of Computer Science, Huddersfield University, Queensgate, Huddersfield HD1 3DH, UK; \\
\textsuperscript{*}Correspondence: U2282847@unimail.hud.ac.uk;
\end{minipage}}

\begin{document}

\maketitle
\begin{abstract}YOLOv4 achieved the best performance on the COCO dataset by combining advanced techniques for regression (bounding box positioning) and classification (object class identification) using the Darknet framework. To enhance accuracy and adaptability, it employs Cross mini-Batch Normalization, Cross-Stage-Partial-connections, Self-Adversarial-Training, and Weighted-Residual-Connections, as well as CIoU loss, Mosaic data augmentation, and DropBlock regularization. With Mosaic augmentation and multi-resolution training, YOLOv4 achieves superior detection in diverse scenarios, attaining 43.5\% AP (in contrast, 65.7\% AP50) on a Tesla V100 at ~65 frames per second, ensuring efficiency, affordability, and adaptability for real-world environments.
\end{abstract}

\keywords{Object Detection; YOLO Models; Convolutional Neural Networks; Real-Time Image processing; Computer Vision; YOLOv4 Framework} 

\section{Introduction}
In recent years, computer vision has emerged as an evolving field, involving machines with the capability to interpret and analyze visual data \cite{RN100,hussain2023custom}. This area enables computers to recognize patterns, process images, and make sense of their surroundings with increasing accuracy. An important element of computer vision is object detection, a sophisticated technique involving the identification and classification of objects \cite{RN102}.

In computer vision, extracting meaningful features from images is essential for accurate analysis and classification. To achieve this, both horizontal and vertical feature extraction techniques were employed using the Histogram of Oriented Gradients (HOG) and the Vertical Histogram of Oriented Gradients (VHOG). These methods allowed for a detailed representation of image structures, capturing variations in gradient orientation to enhance object recognition \cite{RN1}. For classification, machine learning models, i.e., Support Vector Machine (SVM) and Extreme Learning Machine (ELM), were utilized. The SVM approach focused on mapping and visualizing features by localizing key image components and tracking their movement within spatiotemporal regions. However, despite its effectiveness in feature representation, this method encountered limitations in precisely detecting motion-related changes, particularly in dynamic environments. On the other hand, Extreme Learning Machine offered an alternative classification technique, designed to improve computational efficiency and address some of the challenges faced by traditional models \cite{RN2, RN3, RN4,hussain2022feature}. With ongoing advancements in feature extraction and classification methodologies, researchers continue to refine these techniques to enhance accuracy and robustness in various computer vision applications.

The limitations of traditional two-stage object detection approaches, such as prolonged processing times, reliance on complex architectures, and the need for continuous manual adjustments, paved the way for the usage of Convolutional Neural Networks (CNNs) in classification and detection tasks. CNN-based models provided a more efficient and automated approach, significantly enhancing the accuracy and speed of object recognition. A combination of Faster R-CNN and GoogleNet demonstrated superior performance in object detection compared to previous methodologies, highlighting the advantages of deep learning in visual recognition tasks \cite{RN5}. Further advancements were seen in studies employing PoseNet, Mask R-CNN, and MobileNet, which refined image segmentation techniques and boosted overall detection accuracy. Additionally, several other CNN-based architectures, including AlexNet \cite{RN6}, GoogleNet \cite{RN7}, Fast R-CNN \cite{RN11}, ResNet \cite{RN8}, Faster R-CNN \cite{RN12}, Region-based Convolutional Neural Networks (R-CNN) \cite{RN10}, and VGG-Net \cite{RN9}, were instrumental in further improving object identification. These models leveraged deep learning principles to enhance feature extraction, reduce computational overhead, and increase precision in detecting and classifying objects within complex visual environments, paving the way for replacing sensor-based \cite{alsboui2022dynamic} numerical Machine learning focused fault detection in areas such as renewable energy \cite{hussain2022statistical}.

Despite significant advancements in object detection, challenges such as efficient data handling, real-time processing, and the demand for increasingly sophisticated architectures continue to pose obstacles. To address these limitations, the You Only Look Once (YOLO) family of models emerged as a breakthrough, offering a highly efficient alternative to traditional two-stage detection methods. The first version, YOLOv1, was developed using the DarkNet framework and incorporated 24 convolutional layers, while a faster variant, Fast YOLO, featured a reduced architecture with 9 layers to optimize speed \cite{RN14}. This laid the foundation for subsequent enhancements in the YOLO series. YOLOv2, an improved version built on DarkNet-19, introduced key refinements such as batch standardization, optimized bounding box predictions, enhanced class identification, and more crucial feature extraction ~\cite{hussain2022gradient}. Additionally, it integrated size-oriented dimensions, improving its adaptability to objects of varying scales \cite{RN15}. Further evolution led to YOLOv3, which transitioned to the DarkNet-53 architecture, leveraging for more efficient feature extraction. This version also incorporated binary cross-entropy for training optimization, improving classification accuracy while maintaining high processing speed \cite{RN16}. The development of YOLOv4 built upon YOLOv3’s architecture, integrating CSPDarkNet53 as its backbone and PANet with Spatial Pyramid Pooling (SPP) as its neck. These enhancements significantly improved detection accuracy and processing efficiency, making YOLOv4 a powerful tool for real-time object recognition \cite{RN17}.

YOLOv5 achieved remarkable speed and performance by integrating an enhanced CSP-PAN neck with the Spatial Pyramid Pooling-Fast (SPPF) head. These modifications streamlined feature extraction and detection processes, making the model highly efficient for real-time applications \cite{RN18}. Building upon this, YOLOv6 further optimized performance by reducing computational complexity while maintaining high detection accuracy. This made it a preferred choice for applications requiring rapid and precise object recognition without excessive processing demands \cite{RN19}. The introduction of YOLOv7 marked another leap forward, incorporating an auxiliary head and a lead head, significantly enhancing object detection accuracy. This structural refinement allowed the model to better differentiate between objects, improving classification in diverse environments \cite{RN20}. With YOLOv8, the architecture was further enhanced by adopting a modified CSPDarkNet53 backbone and a PAN-FPN neck. These advancements facilitated object identification, image segmentation, and fast movement tracking, a versatile tool for dynamic applications such as surveillance and autonomous navigation. The next-generation YOLOv9 and YOLOv10 models introduced novel techniques to boost detection accuracy even further. YOLOv9 leveraged Programmable Gradient Information and the Generalized Efficient Layer Aggregation Network to refine feature extraction. YOLOv10 built on this by eliminating the need for non-maximum suppression, simplifying post-processing while enhancing performance \cite{RN21, RN22}.

The primary objective of this paper is to conduct an in-depth exploration of YOLOv4, a significant milestone in real-time object detection. YOLOv4 introduced several architectural enhancements that improved detection accuracy, speed, and computational efficiency, making it one of the most robust deep learning models for various applications before introduction of later versions.

\section{Evolution of YOLO}

\subsection{YOLO}

While the YOLO framework transformed object detection with its remarkable speed and efficiency, it was not without its limitations. Due to its structure as grid, early versions of YOLO struggled with detecting smaller objects, particularly when they appeared in clusters. Additionally, the model had difficulty distinguishing complex shapes and positioning. This was largely due to its loss function, which treated errors in small and large bounding boxes equally, leading to inaccuracies in object positioning \cite{RN117, RN28}.

\subsection{YOLOv2}

To address the above-mentioned shortcomings, YOLOv2 introduced significant architectural improvements, adopting the DarkNet-19 framework to enhance feature extraction. The integration of batch normalization improved performance by 2\% mean Average Precision (mAP) while reducing the risk of overfitting through regularization. Furthermore, increasing the input resolution to 448×448 boosted detection accuracy by an additional 4\% mAP, ensuring a more refined and precise object identification process \cite{RN29}. Another crucial advancement in YOLOv2 was the replacement of fully connected layers with anchor boxes, which significantly enhanced recall and improved the ability of the models in detecting objects of varying sizes. The adoption of k-means clustering further optimized the balance between precision and recall by refining anchor box selection, leading to more effective object positioning. These enhancements positioned YOLOv2 as a more robust and accurate object detection model, mitigating many of the challenges faced by its predecessor while maintaining speed and efficiency.

\subsection{YOLOv3}

YOLOv3 improved in object detection by introducing DarkNet-53, a powerful 106-layer deep learning architecture designed for enhanced accuracy and efficiency. This new backbone incorporated residual blocks and upsampling networks, making it larger, more robust, and more precise in detecting objects. One of the key improvements in YOLOv3 was its refined bounding box prediction mechanism, which leveraged a logistic regression model to improve objectness scoring. This enhancement allowed the model to generate more accurate confidence levels for detected objects, reducing false positives and improving detection reliability. Additionally, YOLOv3 abandoned the traditional softmax classification approach in favor of independent logistic classifiers. This shift enabled the model to handle overlapping object categories more effectively, making it better suited for complex scenarios where multiple labels could apply to the same object. Another major breakthrough was YOLOv3’s introduction of three-scale predictions for each location within an input image. This multi-scale detection strategy allowed the model to capture fine-grained semantic details, leading to superior performance in identifying objects of varying sizes \cite{RN30}.

\subsection{YOLOv4}

YOLOv4 introduced a groundbreaking balance between speed and accuracy. With a 10\% increase in detection accuracy and a 12\% boost in processing speed, YOLOv4 set a new standard for real-time object detection. At its core, YOLOv4 utilized CSPDarkNet53, an advanced backbone network with 29 convolutional layers and 3×3 filters, comprising approximately 27.6 million parameters. One of the most notable improvements over YOLOv3 was the inclusion of a Spatial Pyramid Pooling (SPP) block. Furthermore, YOLOv4 replaced YOLOv3’s Feature Pyramid Network (FPN) with the more efficient Path Aggregation Network (PANet), allowing for improved parameter aggregation across different detection layers. It also introduced advanced data augmentation techniques. The mosaic augmentation method allowed the model to train on four different image contexts simultaneously, enhancing its ability to generalize across various object scales and environments. Also, hyperparameters were optimized using genetic algorithms, fine-tuning the model’s performance and robustness across diverse detection tasks \cite{RN31}.

\subsection{YOLOR}

You Only Look One Representation (YOLOR) represents a significant advancement in object detection by unifying multiple learning tasks within a single network. By combining explicit (conscious) and implicit (experiential) knowledge, YOLOR improves its approach to enhancing model performance across various computer vision tasks. The innovation of YOLOR lies in its ability to align features, optimize predictions, and standardize representations for learning different tasks. This integration not only refines the architecture but also improves overall precision by approximately 0.5\%. A key feature of YOLOR is its embedding of implicit representations directly within each Feature Pyramid Network (FPN) feature map, allowing the network to capture and utilize more information from the data. Further refinement is achieved by integrating these implicit representations into the network’s output layers, enhancing the final prediction quality. YOLOR’s approach mitigates the common issue of performance degradation that often arises from the joint optimization of loss functions across multiple tasks. By incorporating canonical representation, YOLOR ensures that each task benefits from shared knowledge without sacrificing accuracy or efficiency. YOLOR achieves superior performance compared to traditional single-task models, offering a robust solution for applications requiring simultaneous object detection, segmentation, and classification. As a result, YOLOR sets a new standard in unified network design, further pushing the boundaries of what is available in real-time computer vision \cite{RN118}.

\subsection{YOLOX}

YOLOX builds upon the DarkNet-53 backbone, refining the YOLOv3 architecture with significant structural and functional improvements. One of its most notable enhancements is the introduction of a decoupled head, which separates the processes of object classification and spatial localization. To enhance training, YOLOX incorporates advanced data augmentation techniques such as Mosaic and MixUp, in order to improve the model’s ability to generalize across various datasets. YOLOX adopts a design that is anchor-free, removing the need for clustering based on anchor. By removing reliance on predefined anchor boxes, the model reduces computational overhead, streamlining the detection process for increased speed and accuracy. A key innovation in YOLOX is SimOTA, a novel label assignment approach that replaces the conventional Intersection over Union (IoU) approach. SimOTA optimizes object matching during training, reducing reliance on manual hyperparameter tuning and significantly lowering training time. This refinement also boosts the model’s mean Average Precision (mAP) by 3\%, further improving detection performance \cite{RN119}.

\subsection{YOLOv5}

YOLOv5 started a major shift in the YOLO family by moving away from the DarkNet framework to PyTorch. This transition provided greater flexibility, ease of deployment, and improved scalability for real-world applications. Retaining CSPDarkNet53 as its backbone, YOLOv5 introduced multiple model sizes, YOLOv5s, YOLOv5m, YOLOv5l, and YOLOv5x, allowing users to select a model that balances accuracy and computational efficiency. A key architectural enhancement in YOLOv5 was the introduction of the Focus layer, which optimized feature extraction by consolidating multiple tasks into a single layer. This not only reduced the overall number of layers and parameters but also significantly improved both forward and backward processing speeds. Importantly, these optimizations were achieved without compromising the model’s mean Average Precision (mAP), ensuring high detection accuracy. YOLOv5 faced challenges in handling overlapping objects. While it incorporated Single Object Tracking (SOT) and Multiple Object Tracking (MOT), both methods encountered tracking errors. However, MOT with dropout tracking demonstrated superior accuracy, mitigating some of these issues and improving object tracking performance \cite{RN32}. Although YOLOv5 has lower accuracy than ResNet and ResNeXt, it excels in multi-category classification by detecting multiple objects within a single frame. Because of its real-time processing power, it excels in applications that require both rapid execution and precise object localization \cite{RN34, RN127, RN129}.

\subsection{YOLOv6}

YOLOv6, crafted using PyTorch, is a high-performance object detection system designed for industrial-grade applications. This iteration introduces several architectural refinements including a hardware-optimized backbone and neck, an improved decoupled head, and a more sophisticated training strategy. These enhancements significantly improve both detection accuracy and computational efficiency, making YOLOv6 a robust choice for real-time object recognition tasks. Results on the COCO dataset highlight YOLOv6’s superior performance in speed and precision. Running on an NVIDIA Tesla, YOLOv6-N achieved an impressive 1234 FPS with an average precision (AP) of 35.9\%. Meanwhile, YOLOv6-S raised the bar with 43.3\% AP at 869 FPS, setting a new standard for efficient detection. Larger variants, YOLOv6-M and YOLOv6-L, further pushed accuracy boundaries, reaching 49.5\% AP and 52.3\% AP, respectively, while maintaining high-speed processing \cite{RN36, geetha2024yolov6}.

\subsection{YOLOv7}

With cutting-edge advancements in both design and training, YOLOv7 sets a new standard for real-time object detection. One of its key innovations is the integration of the Extended Efficient Layer Aggregation Network (E-ELAN), which enhances the model’s ability to capture diverse features, leading to superior performance in object recognition tasks. By incorporating aspects of YOLOv4, Scaled YOLOv4, and YOLO-R, YOLOv7 achieves flexible scaling to match various computational needs. This adaptability ensures optimal performance across different hardware configurations and real-time applications. A major highlight of YOLOv7 is its implementation of the trainable bag-of-freebies approach, which improves detection accuracy without adding to the computational cost of training. This technique optimizes model learning, leading to faster inference speeds and more precise object detection without compromising efficiency \cite{RN37, sundaresan2025performance}.

\subsection{YOLOv8}

YOLOv8 introduces a modular and adaptable architecture, allowing seamless customization and fine-tuning for a variety of computer vision tasks, including object detection, segmentation, and pose estimation \cite{RN38, RN128}. This flexibility makes it the best choice for applications requiring precision and efficiency across diverse scenarios. An advanced development within this framework is YOLO-NAS (Neural Architecture Search)\cite{RN39} , an automated system that optimizes network architecture without the need for manual adjustments. This approach enhances performance while maintaining computational efficiency, striking a balance between high accuracy and low-latency processing. As a result, YOLO-NAS is particularly suited for real-time applications that demand both speed and precision. YOLO-NAS has the ability to dynamically adjust image resolution based on object characteristics, ensuring that computational resources are allocated efficiently during inference. This optimization leads to improved processing speeds while maintaining superior detection accuracy.

\subsection{YOLOv9}

YOLOv9 enables cutting-edge techniques to enhance real-time object detection, focusing on improved training efficiency and feature extraction \cite{RN120}. Programmable Gradient Information optimizes gradient flow during training, making the model to learn more effectively from complex datasets. This ensures faster convergence and improved performance on challenging detection tasks. Another major advancement is the Generalized Efficient Layer Aggregation Network (GELAN), designed to boost both accuracy and processing speed. By enhancing feature extraction and aggregation, YOLOv9 achieves superior precision while maintaining high efficiency, making it best suited for real-time applications. Built upon the Ultralytics YOLOv5 codebase, YOLOv9 sets new benchmarks on the MS COCO dataset, outperforming previous YOLO versions in precision and adaptability across various detection tasks.

\subsection{YOLOv10}

With the removal of non-maximum suppression (NMS) in post-processing, YOLOv10 introduces a game-changing improvement in object detection. This significantly enhances inference speed, making it one of the fastest YOLO models to date. By adopting NMS-free training with dual label assignments, YOLOv10 strikes an optimal balance between speed, accuracy, and computational efficiency. Further architectural improvements include spatial-channel decoupled downsampling, lightweight classification heads, and a rank-guided block design, all of which contribute to reduced computational complexity and a lower parameter count. These optimizations ensure that YOLOv10 remains both scalable and efficient, making it suitable for deployment across a wide range of platforms, from high-performance servers to resource-constrained edge devices. Testing demonstrates that YOLOv9 lags behind YOLOv10 in latency and size of the model, while delivering comparable or superior detection accuracy \cite{RN121}. 

\subsection{YOLOv11}

By introducing groundbreaking architectural enhancements, YOLOv11 achieves faster and more precise object detection. Key advancements include the integration of the C3K2 block, SPFF module, and C2PSA block, each contributing to more efficient feature processing. The C3K2 block employs smaller convolutional kernels to improve computational efficiency, while the SPFF module enhances multi-scale detection, particularly for small objects. Additionally, the C2PSA block incorporates attention mechanisms, enabling the model to focus more effectively on critical regions within an image. The model’s modular backbone, featuring Conv blocks and Bottle Necks, optimizes feature extraction while maintaining processing efficiency. Furthermore, YOLOv11 leverages a multi-scale prediction head, detecting objects across three feature maps to improve precision when identifying objects of different sizes. These enhancements make YOLOv11 a powerful tool for real-time applications requiring high-speed, high-accuracy object detection \cite{RN122}.

\section{Methods} 

\subsection{Bag of freebies}

A Bag of Freebies is a set of techniques designed to improve object detection accuracy without impacting inference costs \cite{RN17, klingler}. A key method is data augmentation, which enhances input variability for robustness. Common approaches include photometric distortions (brightness, contrast, and hue adjustments) and geometric distortions (scaling, cropping, flipping, and rotating). Data augmentation enhances object detection by modifying images while retaining pixel information. Some methods simulate object occlusion, such as Random Erase and CutOut, which replace regions with zeros. Hide-and-Seek and Grid Mask randomly mask multiple areas. Feature-level techniques include DropOut, DropConnect, and DropBlock. Multi-image augmentation methods like MixUp and CutMix blend images and adjust labels accordingly. Style Transfer GAN is also used to reduce texture bias in CNNs. These strategies improve model robustness by increasing image variability, making object detection models more adaptable to different environments without increasing inference cost.

Some Bag of Freebies methods address semantic distribution bias in datasets, particularly class imbalance. In two-stage detectors, this is handled with hard negative example mining or online hard example mining, but these are unsuitable for one-stage detectors due to their dense prediction nature. Instead, Focal Loss was introduced to address class imbalance in one-stage detectors. Another issue is label representation, where one-hot encoding lacks flexibility. Label smoothing converts hard labels into soft labels, improving model robustness. To refine soft labels further, knowledge distillation has been used to enhance training through a label refinement network. The final Bag of Freebies method focuses on Bounding Box regression. Traditional detectors use Mean Square Error (MSE) to regress bounding box coordinates, treating them as independent variables, which ignores object integrity. IoU loss improves this by considering the overlap between predicted and ground truth bounding boxes, making it scale-invariant. Further refinements include GIoU loss, which incorporates shape and orientation, DIoU loss, which adds center distance, and CIoU loss, which combines overlap, center distance, and aspect ratio. CIoU loss achieves better convergence and accuracy, significantly improving Bounding Box regression for object detection models.

\subsection{Bag of specials}

Bag of specials represents additional techniques that slightly affect inference time but result in a marked enhancement of object detection performance. These modules enhance attributes like receptive field, attention mechanisms, and feature integration. Post-processing techniques refine model predictions, ensuring more precise and reliable detection results. Common Bag of Specials modules for enhancing the receptive field include SPP, ASPP, and RFB. Spatial Pyramid Pooling (SPP) enhances Spatial Pyramid Matching (SPM) by replacing bag-of-word features with max-pooling. YOLOv3 improved SPP by concatenating multi-scale max-pooling outputs, boosting AP50 by 2.7\% on MS COCO with only 0.5\% extra computation. ASPP replaces max-pooling with dilated convolutions to expand spatial coverage. RFB (Receptive Field Block) further enhances spatial coverage using multiple dilated convolutions, increasing SSD's AP50 by 5.7\% at just 7\% extra inference time, significantly improving object detection performance.

Attention modules in object detection include channel-wise attention (SE) and point-wise attention (SAM). Squeeze-and-Excitation (SE) improves ResNet50’s top-1 accuracy by 1\% on ImageNet with only 2\% extra computation but increases GPU inference time by 10\%, making it better for mobile devices. Spatial Attention Module (SAM) improves ResNet50-SE by 0.5\% with just 0.1\% extra computation and no impact on GPU inference speed, making it more efficient. Feature integration in object detection started with skip connections and hyper-columns to merge low-level and high-level features. With multi-scale prediction (e.g., FPN), lightweight integration modules emerged, including SFAM (channel-wise re-weighting with SE), ASFF (point-wise re-weighting with softmax), and BiFPN (scale-wise re-weighting with weighted residual connections). Activation functions improve gradient propagation with minimal computational cost. ReLU solved gradient vanishing issues, leading to LReLU, PReLU, ReLU6, SELU, Swish, Hard-Swish, and Mish. These functions enhance model training efficiency and stability, significantly benefiting deep learning models, including those for object detection.

\section{YOLOv4 Architecture}

\begin{figure}[H]
\begin{adjustwidth}{-\extralength}{0cm}
\centering
\includegraphics[width=15cm]{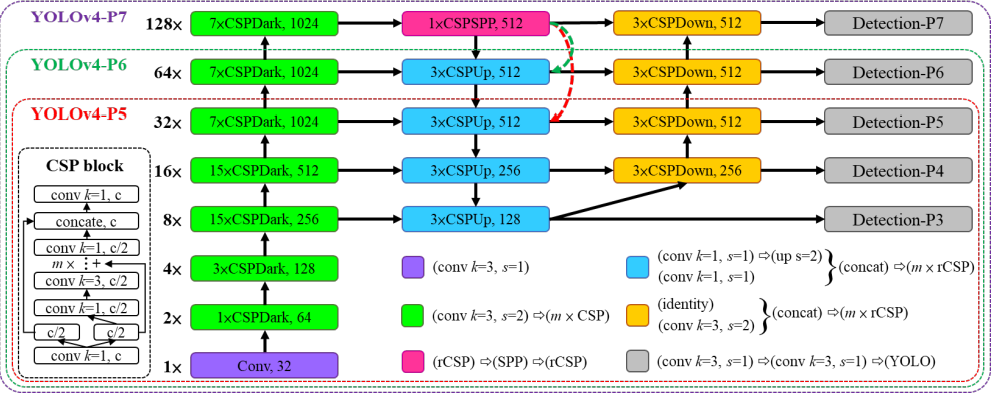}
\end{adjustwidth}
\caption{Architecture of YOLOv4. From \cite{wang2021scaled}}
\label{Figure:1}
\end{figure}

Figure \ref{Figure:1} illustrates the YOLOv4-P5, YOLOv4-P6, and YOLOv4-P7 architectures, detailing the feature extraction, integration, and detection processes. At its core, the CSPDarkNet backbone (green blocks) extracts multi-scale features through hierarchical layers, starting from 1xCSPDark (64 channels) up to 7xCSPDark (1024 channels). Each CSP block (shown separately in the bottom-left corner) consists of multiple convolutional layers, concatenations, and feature map splits, allowing for efficient computation while maintaining accuracy.

To enhance feature fusion, the architecture incorporates PANet (blue blocks), which performs upsampling, concatenation, and convolution operations to merge high-level semantic features with lower-level spatial details. Additionally, the Spatial Pyramid Pooling (SPP) module (pink block) at the highest resolution expands the receptive field, capturing broader contextual information. The CSPDown layers (yellow blocks) are used for feature downsampling, refining extracted features before detection. These layers apply identity connections, downsampling through convolutions (k=3, s=2), and multi-scale feature integration.

The detection process is handled by multi-scale YOLO heads (grey blocks), ranging from Detection-P3 to Detection-P7, depending on the model variant. YOLOv4-P5 supports detection up to P5, while YOLOv4-P6 extends to P6, and YOLOv4-P7 reaches P7 for high-resolution object detection. The final detection layers apply convolution operations and YOLO detection heads for object classification and bounding box regression.

In summary, this architecture balances accuracy, computational efficiency, and scalability, leveraging CSPDarkNet for feature extraction, SPP for receptive field expansion, PANet for feature fusion, and CSPDown for refined multi-scale detection, making YOLOv4 highly effective for real-time object detection across varying object sizes.

The YOLOv4 reveals its important elements. First, the backbone network is responsible for feature formation, which extracts key information from input images. The neck component focuses on feature aggregation, combining features from different layers to enhance detection accuracy. The head handles the critical detection step, identifying and localizing objects within the image. Additionally, YOLOv4 incorporates a "Bag of Freebies," a set of enhancements that improve performance without increasing costs, and a "Bag of Specials," specialized techniques that further boost detection accuracy and robustness  \cite{RN123, RN125, RN126, klingler}.

To achieve better performance without impacting production inference speed, techniques like Bag of Freebies (BoF), including data augmentation using CutMix and Mosaic (with four tiles), regularization of DropBlock, and class label smoothing, were used. By detecting and obscuring the image regions the network heavily depends on training, Self-Adversarial Training (SAT) drives the model to generalize by acquiring new detection features.

\subsection{Backbone}
The backbone acts as the essential classification model that supports the structure of the object detection system and processes the input image by extracting and compressing features using a convolutional neural network. Predictions can be made directly from these backbones, which serve as the network's endpoint, within image classification. By training on ImageNet classification, the backbone network acquires weights optimized for feature recognition, giving object detection systems a head start with a strong foundation. Further refinement of these pretrained weights ensures they meet the specialized needs of object detection in images.

CSPResNext50 and CSPDarknet53, both DenseNet inspired, focus on obstacles caused by gradient loss, refining feature dynamics, encouraging the recycling of features, and simplifying parameter usage, enhancing DenseNet by splitting the base layer features, directing one copy into dense blocks and another forward, effectively cutting bottlenecks and optimizing learning dynamics. EfficientNet, designed by Google Brain, optimizes ConvNet scaling (input size, width, and depth) for image classification, outperforming other networks. In YOLOv4, the neck processes and combines features produced by the ConvNet backbone to prepare for detection. Designs such as FPN, PAN, NAS-FPN, BiFPN, ASFF, and SFAM link the top convolutional layers, improving feature aggregation and flow. By using PANet for feature aggregation and adding an SPP block post-CSPDarknet53, YOLOv4 enhances its receptive field and differentiates the most important aspects from the backbone.

\subsection{Neck}
In YOLO models, the neck serves as a bridge between the backbone and the head, gathering feature maps from multiple stages of the backbone. YOLOv4 incorporates a neck structure that includes an enhanced version of PANet, along with Spatial Pyramid Pooling (SPP) and a Spatial Attention Module (SAM).

Traditional Spatial Pyramid Pooling (SPP) applies fixed-size max pooling to divide feature maps into different regions (e.g., 1×1, 2×2, 4×4), flattening and combining them into a single vector. This loses spatial information, making it better for classification but less effective for object detection. In contrast, YOLOv4's modified SPP retains spatial dimensions by using fixed-size pooling kernels (e.g., 1×1, 5×5, 9×9, 13×13) and concatenating outputs along the channel dimension. This preserves spatial details and expands the receptive field, improving the detection of objects at different scales while maintaining efficiency for real-time performance.

YOLOv3 utilized a Feature Pyramid Network (FPN), but in YOLOv4, this was replaced with a modified version of PANet. Unlike FPN, which follows a top-down approach, PANet introduces an additional bottom-up pathway, enabling lower-level features to be reinforced with contextual information while enriching higher-level features with spatial details. In YOLOv4, this PANet structure was altered to use concatenation instead of traditional addition.

In YOLOv4, the Spatial Attention Module (SAM) was changed by removing average and max pooling, as these operations can reduce valuable feature map information. Instead, the model directly processes feature maps using convolutional layers followed by a sigmoid activation function, generating attention maps that enhance important features while suppressing irrelevant ones, improving object detection accuracy without information loss.

\subsection{Head}
In YOLO models, the head is responsible for object detection and classification. Although YOLOv4 introduced several improvements, it retained the same head as YOLOv3, continuing to generate anchor box predictions and perform bounding box regression. The significant enhancements in YOLOv4’s efficiency and speed are primarily due to optimizations in the backbone and neck rather than the head.

The YOLOv4 architecture is a designed framework for efficient and accurate object detection, balancing computational speed and performance compared to the previous models \cite{RN124} (Table \ref{tab:yolov4_architecture}). It begins with an input image, typically resized to 608 × 608 pixels, which is then normalized. The backbone consists of convolutional layers, starting with a 32-filter 3 × 3 convolution followed by a 64-filter 3 × 3 convolution with downsampling, extracting low-level features from the image. The Cross-Stage Partial (CSP) Blocks are introduced to improve learning efficiency and reduce computational costs by splitting feature maps into two parts, processing one through a series of convolutions, and merging them back. These blocks enhance gradient flow and reduce redundant computations. 

\begin{table}[H]
\caption{YOLOv4 Architecture.\label{tab:yolov4_architecture}}
    \begin{adjustwidth}{-\extralength}{0cm}
        \begin{tabularx}{\fulllength}{CCCCC}
            \toprule
            \textbf{Layer} & \textbf{Filters} & \textbf{Size} & \textbf{Repeat} & \textbf{Output Size} \\
            \midrule
            Input Image    & -                & -             & -               & 608 × 608            \\
            Conv           & 32               & 3 × 3 / 1     & 1               & 608 × 608            \\
            Conv           & 64               & 3 × 3 / 2     & 1               & 304 × 304            \\
            CSPBlock       & 64               & -             & 1               & 304 × 304            \\
            Conv           & 128              & 3 × 3 / 2     & 1               & 152 × 152            \\
            CSPBlock       & 128              & -             & 2               & 152 × 152            \\
            Conv           & 256              & 3 × 3 / 2     & 1               & 76 × 76              \\
            CSPBlock       & 256              & -             & 8               & 76 × 76              \\
            Conv           & 512              & 3 × 3 / 2     & 1               & 38 × 38              \\
            CSPBlock       & 512              & -             & 8               & 38 × 38              \\
            Conv           & 1024             & 3 × 3 / 2     & 1               & 19 × 19              \\
            SPP            & -                & -             & 1               & 19 × 19              \\
            CSPBlock       & 512              & -             & 4               & 19 × 19              \\
            YOLO Head      & 3 Scales         & -             & -               & 19 × 19, 38 × 38, 76 × 76 \\
            \bottomrule
        \end{tabularx}
    \end{adjustwidth}
    \noindent{\footnotesize{*Note: This table summarises the YOLOv4 architecture, featuring CSPBlocks, SPP, and a multi-scale YOLO head for efficient object detection.}}
\end{table}

As the network progresses further, subsequent convolutional layers and CSP blocks operate at increasingly smaller spatial resolutions, achieved through downsampling using stride-2 convolutions.  The architecture includes CSPBlocks with 64, 128, 256, 512, and 1024 filters, repeated multiple times at each stage to capture hierarchical features. At the bottleneck stage, the Spatial Pyramid Pooling (SPP) module aggregates global context by applying pooling at multiple scales, enriching the receptive field. This ensures the network captures both local and global features, which is crucial for detecting objects of varying sizes.

Finally, the YOLO detection head operates at three scales, 19 × 19, 38 × 38, and 76 × 76, to detect objects of different sizes. Each scale utilizes anchor boxes to forecast class probabilities and bounding boxes, ensuring accurate multi-scale detection. The combination of CSP blocks, SPP, and a multi-scale YOLO head makes YOLOv4 highly efficient, optimizing detection accuracy without compromising speed, even for real-time applications.

\subsection{YOLOv4 performance}

Table \ref{tab:yolov4_performance} presents the performance of YOLOv4 on the MS COCO dataset across three different GPU architectures: Maxwell, Pascal, and Volta. The results indicate a trade-off between inference speed (FPS) and accuracy (AP) for different input sizes (416, 512, and 608). For the Maxwell GPU (GTX Titan X / Tesla M40), FPS is significantly lower, with 38 FPS at 416 resolution and 23 FPS at 608 resolution, while the AP gradually increases from 41.2\% to 43.5\% as input size grows. The Pascal GPUs (Titan X, Titan Xp, GTX 1080 Ti, Tesla P100) deliver higher speeds, with 54 FPS for 416 resolution and 33 FPS for 608 resolution, while maintaining similar AP values across sizes. This suggests Pascal GPUs are more efficient at handling YOLOv4 inference compared to Maxwell.

\begin{table}[H]
\centering
\caption{YOLOv4 Performance on MS COCO Dataset across Different GPUs. From \cite{RN17}.\label{tab:yolov4_performance}}
\begin{tabular}{|c|c|c|c|c|c|c|c|c|c|c|}
    \hline
    \textbf{Method} & \textbf{Backbone} & \textbf{Size} & \textbf{FPS} & \textbf{AP} & \textbf{AP$_{50}$} & \textbf{AP$_{75}$} & \textbf{AP$_S$} & \textbf{AP$_M$} & \textbf{AP$_L$} \\
    \hline
    \multicolumn{10}{|c|}{\textbf{Maxwell GPU (GTX Titan X / Tesla M40)}} \\
    \hline
    YOLOv4 & CSPDarknet-53 & 416 & 38 (M) & 41.2\% & 62.8\% & 44.3\% & 20.4\% & 44.4\% & 56.0\% \\
    YOLOv4 & CSPDarknet-53 & 512 & 31 (M) & 43.0\% & 64.9\% & 46.5\% & 24.3\% & 46.1\% & 55.2\% \\
    YOLOv4 & CSPDarknet-53 & 608 & 23 (M) & 43.5\% & 65.7\% & 47.3\% & 26.7\% & 46.7\% & 53.3\% \\
    \hline
    \multicolumn{10}{|c|}{\textbf{Pascal GPU (Titan X / Titan Xp / GTX 1080 Ti / Tesla P100)}} \\
    \hline
    YOLOv4 & CSPDarknet-53 & 416 & 54 (P) & 41.2\% & 62.8\% & 44.3\% & 20.4\% & 44.4\% & 56.0\% \\
    YOLOv4 & CSPDarknet-53 & 512 & 43 (P) & 43.0\% & 64.9\% & 46.5\% & 24.3\% & 46.1\% & 55.2\% \\
    YOLOv4 & CSPDarknet-53 & 608 & 33 (P) & 43.5\% & 65.7\% & 47.3\% & 26.7\% & 46.7\% & 53.3\% \\
    \hline
    \multicolumn{10}{|c|}{\textbf{Volta GPU (Titan V / Tesla V100)}} \\
    \hline
    YOLOv4 & CSPDarknet-53 & 416 & 96 (V) & 41.2\% & 62.8\% & 44.3\% & 20.4\% & 44.4\% & 56.0\% \\
    YOLOv4 & CSPDarknet-53 & 512 & 83 (V) & 43.0\% & 64.9\% & 46.5\% & 24.3\% & 46.1\% & 55.2\% \\
    YOLOv4 & CSPDarknet-53 & 608 & 62 (V) & 43.5\% & 65.7\% & 47.3\% & 26.7\% & 46.7\% & 53.3\% \\
    \hline
\end{tabular}
\end{table}

On the other hand, the Volta GPUs (Titan V / Tesla V100) significantly outperform both Maxwell and Pascal GPUs in terms of FPS, achieving 96 FPS at 416 resolution and 62 FPS at 608 resolution, while preserving the same AP scores as other GPUs. This suggests that Volta GPUs provide the best balance between speed and accuracy, making them the most suitable for real-time object detection applications requiring high throughput. Additionally, across all GPUs, the larger input sizes (608) yield slightly better accuracy (AP 43.5\%) but at the cost of reduced FPS, reaffirming the common trade-off between inference speed and detection precision in deep learning models.

\section{Conclusions}

YOLOv4 introduces a powerful and efficient object detection framework by integrating various enhancements from both the Bag of Freebies (BoF) and Bag of Specials (BoS). BoF techniques, such as data augmentation, label smoothing, knowledge distillation, and advanced bounding box regression losses (IoU, GIoU, DIoU, CIoU), improve model robustness without increasing inference costs. Meanwhile, BoS methods, including SPP, ASPP, RFB, attention modules (SE, SAM), and improved feature integration techniques (SFAM, ASFF, BiFPN), slightly increase computational requirements but significantly boost detection accuracy.

The YOLOv4 architecture, built on CSPDarkNet as the backbone, PANet for feature fusion, and SPP for enhanced receptive field, strikes an optimal balance between speed and accuracy. Multi-scale detection layers, improved activation functions, and refined feature aggregation further strengthen its performance in real-time applications. These innovations ensure that YOLOv4 remains a highly effective solution for object detection across diverse environments while maintaining computational efficiency. The growing reliance on lightweight object detectors is projected to transform automation across multiple fields, including renewable energy ~\cite{hussain2019deployment}, emotional intelligence ~\cite{aydin2023domain,hussain2023child}, and industrial quality inspection.

\begin{adjustwidth}{-\extralength}{0cm}

\bibliographystyle{unsrt}  
\bibliography{ref}  

\end{adjustwidth}
\end{document}